# Smartphone monitoring of smiling as a behavioral proxy of well-being in everyday life

Ming-Zher Poh[1] ✉, Shun Liao[1], Marco Andreetto[1], Daniel McDuff[1], Jonathan Wang[1], Paolo Di Achille[1], Jiang Wu[1], Yun Liu[1], Lawrence Cai[1], Eric Teasley[1], Mark Malhotra[1], Anupam Pathak[1], Shwetak Patel[1]

[1]Google Research

✉ Corresponding author: mingzher@google.com

# Abstract

Subjective well-being is a cornerstone of individual and societal health, yet its scientific measurement has traditionally relied on self-report methods prone to recall bias and high participant burden. This has left a gap in our understanding of well-being as it is expressed in everyday life. We hypothesized that candid smiles captured during natural smartphone interactions could serve as a scalable, objective behavioral correlate of positive affect. To test this, we analyzed 405,448 video clips passively recorded from 233 consented participants over one week. Using a deep learning model to quantify smile intensity, we identified distinct diurnal and daily patterns. Daily patterns of smile intensity across the week showed strong correlation with national survey data on happiness (r=0.92), and diurnal rhythms documented close correspondence with established results from the day reconstruction method (r=0.80). Higher daily mean smile intensity was significantly associated with more physical activity ($\beta$ = 0.043, 95% CI [0.001, 0.085]) and greater light exposure ($\beta$ = 0.038, [0.013, 0.063]), whereas no significant effects were found for smartphone use. These findings suggest that passive smartphone sensing could serve as a powerful, ecologically valid methodology for studying the dynamics of affective behavior and open the door to understanding this behavior at a population scale.

# Introduction

The pursuit of happiness and well-being is a fundamental human endeavor with profound implications for health and longevity. Subjective well-being comprises cognitive evaluations of life satisfaction alongside the experience of positive and negative affect[1]. Recognizing its importance, governments have begun incorporating national happiness indices into policy assessment[2], and researchers have explored mood tracking to promote mental health[3]. However, the scientific study of affect has been constrained by its primary instruments: questionnaires and diaries. These methods, including the day reconstruction method[4] and ecological momentary assessment[5], while foundational, are inherently subjective, susceptible to recall bias, and impose a high participant burden that limits long-term compliance. Consequently, longitudinal studies of well-being in human populations carrying out their ordinary lives are scarce.

An opportunity now exists to create objective, scalable measures of affective states by leveraging ubiquitous personal technology. This approach, known as digital phenotyping[6], involves a paradigm shift from active self-reporting to *in situ* behavioral sensing using data from personal digital devices. We hypothesize that the passive monitoring of spontaneous smiling behavior via a smartphone's front-facing camera provides a reliable and ecologically valid behavioral correlate of positive affect. As a primary expression of positive affect, spontaneous smiling offers a powerful window into subjective well-being. Although people may smile for various reasons, as a smile is a complex social signal, the core connection of smiling to feelings of joy and contentment is fundamental[7]. This link is also bidirectional; feeling good prompts smiles, and the act of smiling can, in turn, facilitate positive feelings[8], making smiling a uniquely valuable and observable behavioral target for study.

In this study, we analyze smile intensity patterns from a diverse group of consented adults during thousands of natural smartphone interactions over a week. We explore the associations between this objective behavioral measure and established determinants of well-being, including day of the week, physical activity, and light exposure. This approach promises to overcome the limitations of traditional methods and facilitate a deeper, more scalable understanding of the dynamics of affective behavior in everyday life.

# Results

**Study Population and Patterns of Behaviors**

We analyzed a large dataset of 405,448 face videos passively recorded using the front-facing smartphone camera from a diverse group of 233 consented US adults [sex: 126 (55.1%) females; age: 38.6 ± 9.8 years, 21.0 to 77.0 years; body mass index (BMI): 28.5 ± 6.9 kg/$m^2$, 18.3 to 61.0 kg/$m^2$; Monk skin tone (MST): 5.6 ± 2.6, 1 to 10 (mean ± SD, range)]. These videos were automatically recorded in the background subsequent to a screen unlock event as participants interacted with their personal smartphones over seven days during their daily lives. Participants uploaded 230.7 ± 172.2 (1 to 1,410) face videos per day. We applied a deep neural network trained to detect smiles on all the videos (details in Methods). The raw smile probabilities generated by the model were highly right-skewed, as is expected for a relatively sparse behavior like spontaneous smiling. We quantified smile intensity as the log-transformed value of the smile probability; this transformation yielded an approximately normal distribution.

Participants also wore a wrist-worn fitness tracker throughout the study period to obtain objective measures of daily step counts and sleep duration. A total of 217 (93.1%) participants yielded 1,387 daily observations with paired tracker data. The mean ± SD (range) minutes wear time was 19.7 ± 2.7 (3.0 to 22.8) hours per day; daily smartphone unlocks was 39.3 ± 26.0 (1.0 to 180.0); daily screen time was 152.2 ± 100.7 (0.4 to 640.9) minutes; daily step count was 8,033.4 ± 4,930.8 (102 to 37,138); daily sleep duration was 6.6 ± 2.0 (0.9 to 14.7) hours; daily light exposure measured using the smartphone ambient light sensor was 474.8 ± 1,228.6 (1.0 to 22,974) lux.

**Weekly and Diurnal Rhythms of Smiling**

We observed a distinct weekly pattern in smile intensity. To account for individual differences, we standardized the smile intensity scores for each person separately[9]. Within individuals, smile intensity was lowest on Monday, rose through the week with a slight dip on Thursday, and peaked on Sunday (Fig. 1A). This objectively and passively measured pattern was strongly correlated with self-reported happiness ($r$ = 0.92, $\rho$ = 0.89) and smiles or laughter ($r$ = 0.93, $\rho$ = 0.88) from a Gallup/Healthways US daily poll[10].

The diurnal rhythm of smile intensity differed based on weekdays and weekends (Fig. 1B). On weekdays, smile intensity dipped in the mid-morning (~9 a.m.) and mid-afternoon (~3 p.m.), corresponding to the typical workday, before rising in the evening. This mid-morning dip was absent on weekends. The diurnal pattern of positive affect showed strong correlation ($r$ = 0.80, $\rho$ = 0.88) with previous results using the day reconstruction method[11].

**Associations between Smiling and Health Behaviors**

To explore relationships between daily smiling and modifiable behaviors, we used a linear mixed-effects model adjusted for multiple covariates including age, sex, BMI, wear duration of fitness tracker, and weekends (Fig. 1C). Higher daily step counts ($\beta$ = 0.043, 95% CI [0.001, 0.085]; $P$ = 0.044) and greater ambient light exposure ($\beta$ = 0.038, [0.013, 0.063]; $P$ = 0.003) were significantly associated with increased smile intensity. Contextual factors also played a role, with smile intensity being higher on weekends ($\beta$ = 0.071, [0.020, 0.122]; $P$ = 0.026$) and for female participants ($\beta$ = 0.274, [0.033, 0.516]; $P$ = 0.026). No significant effects were found for age, BMI, sleep duration, number of smartphone unlocks or screen time.

A key finding relates to the heterogeneity of these effects. The model included significant variance in baseline smile intensity between individuals (intercept variance = 0.763), indicating that some individuals have a higher smile intensity than others on average. We also found variance in the slope for total steps (slope variance = 0.012), indicating that the strength of the relationship between physical activity and smile intensity differs substantially across individuals. The inclusion of this random slope showed a significant improvement in model fit over a random-intercept-only model (▯$^2$ = 6.725; $P$ = 0.035).

For a more intuitive metric of effect size, we compared, within each person, the smile intensity on their highest versus lowest behavioral expression days using a bootstrapped repeated-measures Cohen's $d_z$. A day with at least 1,000 more steps was associated with a modest positive effect on smile intensity ($d_z$ = 0.235, [0.048, 0.257]). Similarly, a day with high light exposure (≥500 lux increase) had a positive effect ($d_z$ = 0.191, [0.053, 0.340]).

# Discussion

In this study, we introduced and validated a novel method for objectively measuring a behavioral correlate of positive affect—spontaneous smiling—passively and at scale in a real-world setting. By analyzing over 400,000 video snippets from everyday smartphone use, we demonstrated that this technique can reveal robust weekly and daily rhythms of affective expression that align remarkably well with established, survey-based measures of well-being. Unlike traditional studies that rely heavily on self-report, which is susceptible to recall and social desirability biases, our approach captured spontaneous smiling behavior alongside passively sensed data on physical activity, light exposure, sleep, and screen time.

We found that affect, as measured by smile intensity, is not static but follows clear temporal patterns. The peaking of smile intensity from Friday to Sunday and the diurnal dips during weekday work hours are consistent with the well-documented ``weekend effect''[10] and provide further evidence for the chronobiology of positive affect and its entrainment to work schedules. The strong correlations with large-scale national poll data[10] and established results from the day reconstruction[11] method provide external validation for our method, suggesting that passive sensing can capture meaningful population-level trends.

Our results both align with and extend prior work. The positive association between smile intensity and both physical activity[12] and light exposure[13] is consistent with prior literature on determinants of well-being. Our finding that the effect of physical activity is person-specific suggests that interventions aiming to improve well-being may be more successful if they are tailored to an individual's unique response profile. The lack of association between phone unlocks and screen time with smile intensity aligns with the negligible impact of smartphone use on mood and well-being observed in a large-scale US population study[14].

It is important to distinguish the data collection protocols mandated by this study's ethical framework from the envisioned real-world deployment of the technology. While we required participants to manually review and approve video clips to ensure strict privacy standards during this validation study, this active user intervention is not an inherent requirement of the methodology itself. The core approach is designed to be entirely passive, capable of being implemented as an on-device, real-time process that extracts smile metrics without permanently storing or transmitting raw video footage. By processing data locally, the method preserves user privacy by design while removing the need for manual data management. Consequently, this passive monitoring contrasts sharply with traditional subjective measures like ecological momentary assessments or the day reconstruction method, which impose significant participant burden through frequent, interruptive surveys, and additionally primes and potentially biases participant responses. Because it requires no additional effort beyond normal smartphone interaction, the proposed approach offers a uniquely scalable and frictionless solution for longitudinal monitoring.

The data utilized in this study were derived from a retrospective analysis of a dataset originally collected to validate passive heart rate monitoring via smartphone use with consent for general research and development[17]. This incidental data capture offered a distinct methodological advantage regarding the authenticity of the observed behaviors. Because the original study focused on physiological sensing, participants were entirely naive to the specific objective of analyzing facial expressions. This lack of awareness minimized the potential for demand characteristics or the Hawthorne effect, where participants might otherwise subconsciously alter their expressions if they knew their emotional displays were being scrutinized. Thus, this retrospective analysis is based on truly spontaneous and naturalistic expressions, free from the performance biases that may be introduced by explicit affect monitoring studies.

Our study has its limitations. A smile is a complex signal and not a perfect proxy for happiness; our model quantifies smile intensity without explicitly classifying Duchenne versus non-Duchenne smiles, the former being a key indicator of genuine enjoyment[15]. However, our aim was to capture a continuous behavioral correlate of positive affect, an approach grounded in research suggesting the Duchenne marker, activation of the orbicularis oculi (cheek raiser) muscle, may be an artifact of overall smile intensity[16]. As smiling is a sparse event, a continuous intensity metric provides a richer and more statistically powerful signal for detecting the subtle temporal rhythms central to our findings than a binary classification would allow. Crucially, the context of data collection—private interactions with a personal device captured passively, rather than with a social audience—also reduced the likelihood of purely social or posed smiles, strengthening the interpretation of the captured expressions as spontaneous. Future work could decode specific facial action units to further disentangle the roles of smile type and intensity. Our study lacks the context of the smartphone interaction; parallel analysis of on-screen content is an important next step to disentangle the triggers of affective responses.

Further limitations concern the study's generalizability and broader societal implications. Our findings are based on a US sample and may not extend to cultures with different social display norms. While promising for public health, such technologies carry dual-use risks, including the potential for misuse in surveillance or manipulation. It is therefore imperative that technical safeguards are paired with the development of robust ethical guidelines and regulatory frameworks to govern their responsible deployment.

This study demonstrates that passive sensing of smiling behavior via smartphones is a viable and powerful new instrument for the study of well-being. This approach moves the measurement of affect out of the lab and into the fabric of everyday life, providing a scalable, objective, and continuous window into the affective rhythms of individuals and populations. By overcoming the limitations of traditional methods, this technology has the potential to transform our understanding of mental and emotional health, paving the way for real-time monitoring of societal well-being and the development of personalized interventions to help individuals lead happier, healthier lives.

# Methods

## Study Details

This study was approved by Advarra Institutional Review Board, Columbia, MD. We obtained informed consent from all participants, whose participation was entirely voluntary, and the study was conducted in accordance with the principles of the Declaration of Helsinki. To mitigate the inherent risk of identification associated with collecting face videos, several procedural and management protections were implemented. See SI Appendix for details.

Before any study-related assessments and procedures were performed, study participants were required to read and sign an informed consent form which outlined the study procedures and explicitly acknowledged the inherent risk of loss of privacy and re-identification due to the collection of pictures and videos of their face. Consented study participants installed custom software on their personal smartphone and then used their phone throughout the day as they normally would. Brief video clips were periodically recorded using the front-facing camera when the phone was in use. While study participants were aware of general research and development, they were naive to the specific aim of smile analysis, minimizing the potential for observer effects. To counter the risk of identification and incidental data capture, multiple technical and procedural checkpoints were implemented to safeguard the inherently identifiable face videos. The safeguards implemented included participant review of video clips, restrictions on recording conditions, and cropping of videos to only show the participant’s face. Recording could only occur within 10 minutes of the phone being unlocked and only while a face was visible. Recording immediately stopped whenever the phone was locked or if no face was in frame. This was intended to reduce the chance of recording someone other than the participant. Videos were cropped automatically on device to only the participant’s face to avoid recording of the background environment. The front-facing camera indicator light on the phone was activated whenever a video recording was in progress. Video data was stored locally and never left the phone until it was manually reviewed and approved for upload by the participant. Participants were explicitly instructed to view each video clip and delete clips containing other people or any sensitive subjects. Study participants could choose to delete any video face recordings and only upload video files that they were comfortable sharing with Google for study purposes. The majority of videos were approved by participants (mode = 95%, median = 84.4%, IQR = 22.9%). Once uploaded, information that could personally identify the participant (such as name or email) was separated from the study data and replaced with a unique Subject ID. Although the face videos remain inherently identifiable, study participants were informed this data will not be shared with affiliates or research partners, and are stored securely. Participants also wore a wrist-based fitness tracker, a Fitbit Charge 6 (Google, Mountain View, CA), throughout the study to obtain objective measures of daily step counts and sleep duration.

## Video Specifications

Videos were fixed to a duration of 8 seconds and recorded at VGA (640 x 480) resolution, 15 FPS, and with the phone's default 3A settings (autoexposure, autofocus, auto-white balance) enabled, and saved as Motion JPEG (M-JPEG) at maximum quality to avoid inter-frame compression. Clips shorter than 8 seconds were discarded on device. For each screen unlock, recording would repeat for a cumulative maximum of 10 minutes.

## Deep Learning Model for Smile Analysis

We used a deep learning model with a MobileNet architecture trained to infer various face attributes, including smiling[18]. To ensure good labeling replication, three human raters were used to annotate each face in the training set containing 130,451 images (47.1% with smiles). For the test set, five human raters were used to annotate 10,721 images (41.4% with smiles). The overall accuracy of the deep learning model for smile classification was 90.8%. Accuracy was consistently high for males (90.6%) and females (91.2%), across age buckets of <20 years (90.6%), 20-60 years (90.9%), 60+ years (91.0%), and across skin tone groups of light (90.9%), medium (91.6%), dark (91.3%).

## Objective Measurements

We applied the deep learning model above to all the available valid video frames to detect smiles. Valid video frames were those that contained a human face with the mouth visible, without face coverings (e.g. masks), and were not under exposed or blurred. We quantified smile intensity as the log-transformed value of the model-predicted smile probability. For each video, we computed the mean smile intensity across all valid frames. To account for individual differences in smile intensity, we standardized the smile intensity scores for each person separately by subtracting their average smile intensity from their score, following the procedure described by Golder and Macy[9]. We computed the number of smartphone unlocks and screen time per participant from Android system event logs. To remove outliers, session durations were filtered by the 99th percentile of session durations observed in a large US adult population[14]. Light exposure was measured using the smartphone's ambient light sensor.

## Statistical Analysis

To account for the nested structure of the data (daily observations within participants), we fit a linear mixed-effects model using the statsmodels package (v0.14.0) in Python. The outcome variable was the normalized smile intensity. The model included fixed effects for participant-level demographic variables (age, sex, BMI) and daily time-varying predictors (weekend, total steps, ambient light exposure, screen time, sleep duration, and tracker wear time). All continuous predictors were z-score normalized. The final random effects structure was determined by sequentially comparing nested models via likelihood ratio tests. The final model included random intercepts for each participant to account for baseline differences in well-being and a random slope for daily total steps to test for individual differences in its association with the outcome. To provide an intuitive, standardized measure of effect size, we conducted a non-parametric within-subject analysis. For each key behavior, we calculated the repeated-measures Cohen's $d_z$ by taking the difference in well-being score on the day of maximum behavioral expression and the day of minimum expression and dividing by the standard deviation of these differences. This analysis was restricted to participants who showed sufficient variation in the behavior (a ≥1000 difference for step counts and a ≥500 lux difference for light exposure between min/max days). A total of 207 (95.4%) and 130 (59.9%) participants met this criteria for steps and light exposure, respectively. To derive stable 95% confidence intervals, this procedure was repeated on 10,000 bootstrapped resamples of the data.

## Acknowledgements

We thank Tracy Giest, Jonathan Hsu, Sam Mravca, Alex Mun, Derrick Vickers, Brent Winslow, Xiaoxia Zhang for their assistance in data collection.

## Competing interests

This study was funded by Alphabet Inc and/or a subsidiary thereof (‘Alphabet’). All the authors are employees of Alphabet and may own stock as part of the standard compensation.

## Data Availability

A CSV file containing de-identified daily aggregated data is provided in the SI Appendix to facilitate reproduction of the statistical analysis for this research. This dataset contains the day of week, daily smile intensity, total steps, light exposure, sleep duration, tracker wear time, screen time, number of unlocks per participant along with their age, sex, and body mass index.

## Code Availability

A pseudocode implementation of the algorithms is available in Supplementary Information.

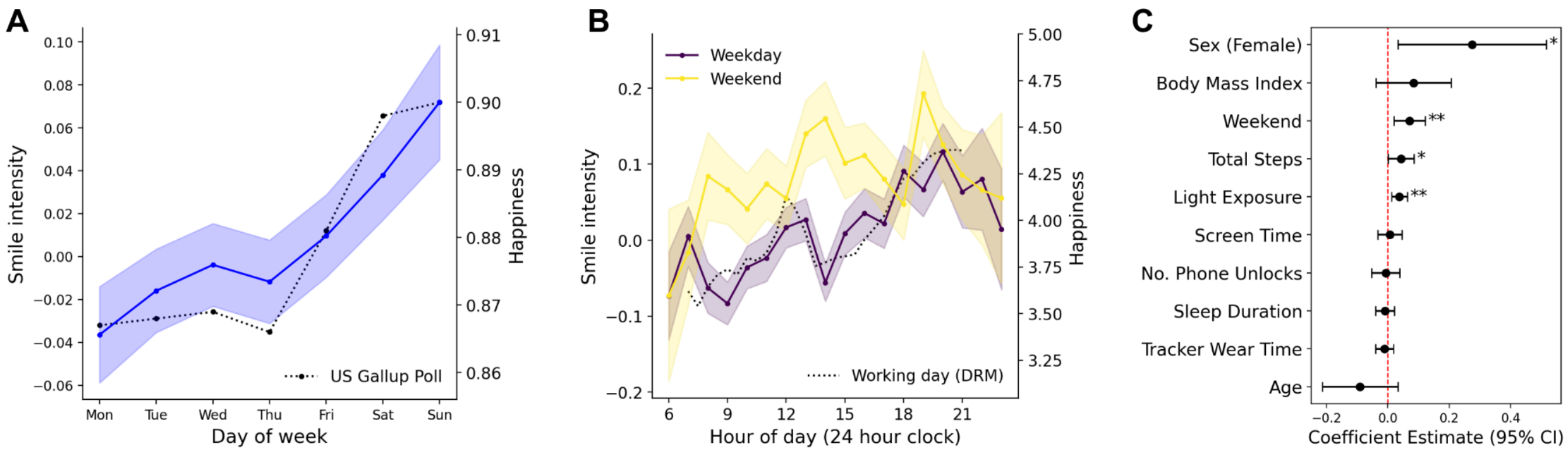

**Figure 1 |** (A) Daily changes in individual smile intensities. (B) Hourly changes in individual smile intensity aggregated by weekdays versus weekend. Each solid trace shows the mean smile intensity and the shaded region corresponds to the standard error of measurement. (C) Associations between daily behaviors and smile intensity from the linear mixed-effects model. The forest plot displays the fixed-effect coefficient estimates ($\beta$) as points and their corresponding 95% confidence intervals as horizontal lines. Coefficients are interpreted as the change in the normalized log-smile score for a one standard deviation increase in the predictor variable, adjusted for all other variables in the model. The vertical dashed line indicates no effect. Asterisks denote statistical significance: *$P$ < 0.05; **$P$ < 0.01.